\documentclass[journal]{IEEEtran}
 \pdfoutput=1
\usepackage[normalem]{ulem}
\usepackage{bm}
\usepackage{amsmath,amssymb,amsfonts}
\usepackage[c1]{optidef}
\usepackage{comment}
\usepackage{hyperref}
\usepackage{booktabs}
\usepackage{makecell}
\usepackage{xcolor}
\usepackage[noadjust]{cite}
\usepackage{algorithm, algorithmic}

\usepackage[caption=false,font=footnotesize]{subfig}
\usepackage{tabularx}
\usepackage[normalem]{ulem}
\usepackage{comment}
\usepackage{xcolor, soul}
\usepackage{tikz}
\usetikzlibrary{shapes,arrows}
\usepackage{smartdiagram}

\newcommand{\black}[1]{\textcolor{black}{#1}}

\newcommand{\Vb}{\mathbf{b}}
\newcommand{\Vw}{\mathbf{w}}
\newcommand{\Vx}{\mathbf{x}}

\newcommand{\Vtheta}{\bm{\theta}}

\newcommand{\Valpha}{\bm{\alpha}}

\newcommand{\matW}{\mathbf{W}}

\newcommand{\matD}{\mathbf{D}}

\DeclareMathOperator*{\argmin}{arg\,min} 
\DeclareMathOperator*{\argmax}{\arg\,\max}

\begin{document}

\title{Learning Data-Driven Uncertainty Set Partitions \\
for Robust and Adaptive Energy Forecasting \\ with Missing Data}

\author{Akylas~Stratigakos,~\IEEEmembership{Member,~IEEE,} and~Panagiotis~Andrianesis,~\IEEEmembership{Member,~IEEE}
        
\thanks{This work was supported in part by the Leverhulme Trust International Professorship (LIP-2020-002) and in part by the European Research Council (ERC) Starting Grant CRISP under Agreement 101165616.}
\thanks{A. Stratigakos is with the Department of Electrical and Electronic Engineering, Imperial College London, London, SW7 2AZ, U.K.: a.stratigakos@imperial.ac.uk.
P. Andrianesis is with the Center for Processes, Renewable Energy and Energy Systems (PERSEE), Mines Paris, PSL University, 06904 Sophia Antipolis, France: panagiotis.andrianesis@minesparis.psl.eu.}
}        
\maketitle

\begin{abstract}
Short-term forecasting models typically assume the availability of input data (features) when they are deployed and in use.
However, equipment failures, disruptions, cyberattacks, may lead to missing features when such models are used operationally, 
which could negatively affect forecast accuracy, 
and result in suboptimal operational decisions.
In this paper, we use adaptive robust optimization and adversarial machine learning to develop forecasting models that seamlessly handle missing data operationally.
We propose linear- and neural network-based forecasting models with parameters that adapt to available features, 
combining linear adaptation with a novel algorithm for learning data-driven uncertainty set partitions.
The proposed adaptive models do not rely on identifying historical missing data patterns and are suitable for real-time operations under stringent time constraints. 
Extensive numerical experiments on short-term wind power forecasting considering horizons from 15 minutes to 4 hours ahead illustrate that our proposed adaptive models are on par with imputation when data are missing for very short periods (e.g., when only the latest measurement is missing) whereas they significantly outperform imputation when data are missing for longer periods. 
We further provide insights by showcasing how linear adaptation and data-driven partitions (even with a few subsets) approach the performance of the optimal, yet impractical, method of retraining for every possible realization of missing data.
\end{abstract}


\IEEEpeerreviewmaketitle

\section{Introduction}

Variable renewable energy sources, 
such as wind and solar, 
dominate low-carbon power systems.
To deal with their inherent uncertainty and variability, 
system operators manage operational risk 
based on a forward-looking grid status estimation \cite{crof_gpt}.
For instance, they run short-term scheduling applications 
to evaluate the reliability of market-based dispatch, 
which are based on short-term energy forecasts 
with a horizon ranging from a few minutes to several hours ahead \cite{helman2008design}.

\subsection{Background and Motivation}

A critical assumption underpinning the forecasting models is that input data, \emph{a.k.a.} features, 
such as weather forecasts and production measurements, 
will always be available when needed,
i.e., when the models are deployed to produce ``operational'' forecasts.
However, equipment failures, disruptions, cyberattacks, may lead to missing data 
when the forecasting model is deployed \cite{polyzotis2017data},
thus compromising forecast accuracy 
and leading to suboptimal operational decisions \cite{COVILLE20114505}.
For example,
\black{weather data typically used as input to short-term energy forecasting are} sometimes published with a delay \cite{mars-survey}; 
similarly, some production measurements may not be available in real-time \black{but retrieved ex-post by system operators}.
As a result, forecasters are often provided with complete data sets for model training and offline testing,
but in real time, when these models need the (same) features to produce forecasts operationally,
some of them may be unavailable. 
Hence, robustness against missing data in real time is critical.
Perhaps most importantly, 
because such models support near real-time processes, 
\black{such as short-term scheduling applications,}
dealing with missing data should comply
with stringent time constraints.

\subsection{Related Work}

So far, missing data are handled with imputation \cite{rubin1976inference}, i.e., a missing value is replaced with a single or multiple ``plausible'' values \cite{white2011multiple}.
The imputation method can be simple (e.g., single mean imputation) or based on an iterative process (e.g., multiple imputation).
When the training data set contains observations with missing data, imputation is applied
both before model training and when the model is used operationally.
In such a case, single imputation methods (e.g., mean imputation) may suffice for consistent forecasts \cite{bertsimas2024simpleimputation}; 
however, they may rely on ad hoc, potentially hard-to-verify, assumptions about the missingness mechanism.
\black{Learning-based imputation methods \cite{cao2018brits} address this issue by directly learning from observed missing data.}
\black{However, historical observations with missing data may not be available;
this case is of particular interest in power system applications where inputs are sometimes made available with delay (e.g., weather data \cite{mars-survey}) or missing inputs (e.g., production measurements) are retrieved ex-post.} 
In this more challenging case, which is our main interest, 
imputation methods can only rely on iterative approaches,
as single imputation introduces forecast bias \cite{josse2024consistency}.
Alternatively, a new model can be retrained, 
either on the fly or offline, 
considering only the available features;
retraining is shown to be asymptotically optimal \cite{josse2024consistency}.
However, both can soon become impractical.
Iterative imputation requires additional training 
after the missing data are realized, 
which becomes computationally prohibitive 
for real-time processes, 
whereas retraining requires an additional model for each of the missing feature combinations, 
whose number grows exponentially with the number of features.
To address these shortcomings, the pertinent idea is to avoid imputation altogether by embedding missing data within the forecasting model.
An insightful recent work \cite{bertsimas2024adaptive} develops a hierarchy of regression models, placed between mean imputation and retraining, with parameters that adapt to the available features.

The application domain for this paper pertains to energy forecasting.
In particular, we focus on short-term wind power forecasting with a horizon of up to several hours ahead,
which is a critical process for operational reliability.
In this process, 
\black{the industry standard is imputation with persistence forecasts, 
i.e., forward-fill the last measured value over the next periods, 
which performs well at a minimal computational cost.}
In \cite{tawn2020missing}, several imputation methods are compared with retraining without missing features; 
retraining performs best.
Considering historical observations with missing data, 
\cite{pierrot2025data} imputes missing features using spatial information across adjacent wind plants,
whereas \cite{wen2024probabilistic} develops neural network-based models with layer biases that adapt to available features.
\black{In \cite{wen2023wind, wen2024tackling}, 
a probabilistic approach that jointly considers imputation and forecasting is implemented, based on generative models \cite{yoon2018gain}.
However, estimating missing features and deriving forecasts requires implementing a costly sampling procedure when the models are used operationally, 
which may be unsuitable for real-time processes.}

To deal with missing data in energy forecasting applications with complete training data sets, 
\cite{stratigakos2023towards} proposes a robust regression that minimizes the worst-case loss due to missing features, for linear models and piecewise linear loss functions.
Adversarial learning \cite{madry2018towards} is employed in \cite{wangkun2023} for neural network-based load forecasting using a robust formulation;
however, this requires expressing neural network training as a mixed-integer problem that is repeatedly solved, which may render it computationally prohibitive.

A critical design component of robust formulations \cite{bertsimas_hertog2020} is the \emph{uncertainty set} 
that captures possible uncertainty realizations --- see, e.g., \cite{PESGM23, PESGM24, Bertsim2025} for applications of constructing uncertainty sets for wind power forecasting.
Large uncertainty sets may lead to overly conservative model parameters.
Finite adaptability \cite{bertsimas2010finite}, 
introduced in the context of adaptive optimization problems with discrete decisions,
reduces conservativeness by partitioning the uncertainty set into smaller, less conservative, subsets. 
Finding a good partition is hard; 
\cite{postek2016multistage} proposes several partitioning methods for adaptive problems with discrete decisions, 
whereas \cite{stratigakos2023interpretable} develops a tree-based learning approach for data-driven partitions.

\subsection{Aim and Contribution}

In this paper, we develop short-term wind power forecasting models that seamlessly adapt to available features when the models are deployed and used operationally.
Inspired by \cite{bertsimas2024adaptive}, 
we consider linear- and neural network-based forecasting models with adaptive parameters. 
We implement adversarial learning by formulating model training as an adaptive robust optimization problem;
namely, we combine linear adaptation with uncertainty set partitioning and minimize the worst-case loss due to missing features.
We examine the impact of linear adaptation, compare data-driven to ad-hoc partitioning (implemented in \cite{wangkun2023,stratigakos2023towards}), 
and assess our methods against standard industry imputation approaches.
Notably, our methods do not rely on identifying historical missing data patterns,
and are suitable for deployment in real-time operations under stringent time constraints.

Our main contribution is three-fold.
Firstly, we propose a tailored gradient-based strategy to train robust and adaptive forecasting models by rapidly finding adversarial missing feature examples.
Secondly, we develop a novel tree-based algorithm 
to learn data-driven uncertainty set partitions.
We posit that a small number of missing feature combinations is critical to forecast accuracy when data are missing; 
our partitioning algorithm finds such combinations and, thus, approaches the optimal, yet impractical, retraining method.
Thirdly, we provide insights through extensive numerical experimentation using real-world data on short-term wind power forecasting.
The results indicate that when data are missing for very short periods (e.g., the latest wind power measurement is missing), our methods are on par with imputation;
when data are missing for longer periods, 
our methods perform significantly better,
hedging effectively against worst-case scenarios.

\subsection{Paper Structure}

The rest of the paper is organized as follows.
First, we formulate adaptive forecasting models (Section~\ref{adapt_models_section}) and develop a partition learning methodology (Section~\ref{partitioning_section}).
Next, we detail our experimental setup (Section~\ref{exp_setup_section}) and present the results of the numerical experiments (Section~\ref{results_section}).
Finally, we conclude and provide directions for future work (Section~\ref{conclusions_section}).

\section{Training with Missing Data} \label{adapt_models_section}

In this section, 
we present preliminaries on forecasting models with missing data 
using uncertainty sets (Subsection \ref{prelim_subsection}), 
we formulate robust and adaptive robust forecasting models (Subsection \ref{reg_subsection}), 
and we describe a training process with adversarial examples (Subsection \ref{adv_learn_methodology_subsection}).

\subsection{Model Preliminaries}\label{prelim_subsection}

\paragraph*{Notation}
We use bold font lowercase (uppercase) for vectors (matrices), 
calligraphic for sets,
and $|\cdot|$ to denote set cardinality.
Consider a forecasting task with $y \in \mathcal{Y}$ being the target variable 
and $\Vx \in \mathcal{X}$ a $p$-size feature vector.
For example, in short-term wind power forecasting, 
$y$ is the production of the $s$-th plant at period $t+h$ 
($t$ is the time index and $h$ the forecast horizon), 
whereas $\Vx$ typically concatenates the historical production 
across adjacent farms for periods $[t, t-\tau]$  
($\tau$ is the maximum historical lag) 
and other relevant information such as the weather. 
We assume access to an $n$-sized training data set 
without any missing data, 
denoted by 
$\mathcal{D}=\{(y_i, \Vx_i)\}_{i\in[n]}$, 
where $[n] = \{1, \dots, n\}$.

In a standard regression setting, 
we train a forecasting model $f: \mathcal{X} \rightarrow \mathcal{Y}$, 
parameterized by $\Vtheta$, 
by minimizing the empirical loss over $\mathcal{D}$, 
given by
\begin{equation}\label{emp_loss}
{L}_\mathcal{D}\big(y,f(\Vx;\Vtheta)\big) = \frac{1}{n}\sum\limits_{i\in [n]}l\big(y_i, f(\Vx_i;\Vtheta)\big), 
\end{equation}
where $l: \mathbb{R}\times\mathbb{R}\rightarrow \mathbb{R}_+$ is a convex loss function, e.g., the mean squared error or the quantile loss, 
whereas $y$ and $\Vx$ refer (with some abuse of notation) 
to the items of data set $\mathcal{D}$.

\subsubsection{Base Forecasting Models}
We focus on gradient-based forecasting models, which comprise two widely used classes of models, 
linear regression and neural networks, 
hereinafter referred to as base forecasting models.
\paragraph{Linear Regression (LR)} We consider an LR base forecasting model given by
\begin{equation} \label{LR-base}
   f(\Vx;\Vtheta) = \Vw^{\top}\Vx + b, 
\end{equation}
where $\Vw$ are linear weights, $b$ is the bias, and $\Vtheta := (\Vw, b)$. 
\paragraph{Neural Network (NN)} We consider an $M$-layer feed-forward NN  base forecasting model with a rectified linear unit activation function, given by
\begin{subequations} \label{NN-base}
\begin{align}
    & \mathbf{g}^1 = \matW^{0} \Vx + \Vb^{0}, & \\
    & \mathbf{g}^{m+1} = \max(\matW^{m} \mathbf{g}^{m}  + \Vb^{m}, \mathbf{0}), & \,\, 1 \leq m < M,\\    
    & {f}(\Vx; \Vtheta) = \Vw^{\top}\mathbf{g}^{M}  + b, & 
\end{align}
\end{subequations}
where $\matW^m$ and $\Vb^m$ are the appropriately-sized weights and bias, respectively,
of the $m$-th hidden layer, 
$\Vw$ and $b$ are the weights and bias, respectively, 
of the output linear layer, 
and $\Vtheta : =\{(\matW^{0:M-1}, \Vb^{0:M-1}), (\Vw, b)\}$.

\subsubsection{Modeling Missing Data}\label{modeling_missing_subsection}
In our setting, missing data pertain to feature availability.
Although we assume that all features are available 
when we train the forecasting model,
at the time the model is deployed 
(and used to produce operational forecasts), 
some features may not be available.
Thus, we model the availability of the $j$-th feature 
by introducing binary variable $\alpha_j$,
which equals $1$ when the feature is missing,
and $0$ otherwise.
Let $\mathcal{P}$ denote the set of features that may not be available 
(indeed, \black{some features are always available, e.g., calendar variables, and not included in $\mathcal{P}$)}.
Feature availability is thus reflected in the feature vector as follows:
\begin{equation}\label{x-alpha}
{\Vx}({\Valpha})=
\begin{cases}
    x_j (1 - \alpha_j),& \forall j \in \mathcal{P},\\
    x_j,              & \text{otherwise}.
\end{cases}    
\end{equation}
We note that \eqref{x-alpha} replaces a missing value of $x_j$ with zero as a placeholder.
For a given realization of missing features, say $\hat{\Valpha}$,
we derive a respective set of parameters $\hat\Vtheta$, given by
\begin{equation}\label{nominal_a_model}
   \hat{\Vtheta} \in {\argmin}_{\Vtheta}~
{L}_\mathcal{D}\big(y,f(\Vx(\hat{\Valpha});\Vtheta)\big),
\end{equation}
where $\Vx_i({\hat{\Valpha}})$ is defined by \eqref{x-alpha}
by applying (the same) $\Valpha$ to all observations.

Algorithm~\ref{nom_train_algo} presents
a batch gradient descent approach to solve \eqref{nominal_a_model},
which requires controlling for a set of hyperparameters (learning rate $\eta$, maximum number of iterations $K$, and patience threshold $\Phi$).
It starts by creating training/validation subsets, 
without shuffling, 
such that $\mathcal{D}=\mathcal{D}^{\text{train}}\cup\mathcal{D}^{\text{val}}$ and randomly initializing base forecasting model parameters $\hat\Vtheta$ (lines~\ref{alg1_step1} and \ref{alg1_step2}).
At iteration $k$, 
a training-loss-based gradient update is performed (line~\ref{step_grad_update}),
and the performance is evaluated by monitoring the validation loss (line~\ref{step_validation_loss}). 
The algorithm terminates (line~\ref{while_loop_1}) 
when the maximum number of iterations, $K$, is reached
or the validation loss does not improve for $\Phi$ consecutive iterations (monitored by counter $\phi$ in lines~\ref{algo1_step6}-\ref{algo1_line10}). 
It returns parameters $\hat\Vtheta$, 
and the resulting validation loss, $\hat{L}$.
Implementing a mini-batch version of Algorithm~\ref{nom_train_algo} with an adaptive learning rate is left to the interested reader.

\begin{algorithm}[tb!]
\caption{ \texttt{Training Process}}\label{nom_train_algo}
  \textbf{Input:} Data set 
  $\mathcal{D}$, missing feature realization $\hat\Valpha$, 
  learning rate $\eta$, number of iterations $K$, patience threshold $\Phi$.
  \\
  \textbf{Output:} Parameters $\hat\Vtheta$, loss $\hat{L}$.
  \begin{algorithmic}[1]
  \STATE \label{alg1_step1} 
  Set $k = 0$,  $\phi=0$. 
  Split $\mathcal{D}=\mathcal{D}^{\text{train}}\cup\mathcal{D}^{\text{val}}$.
\STATE \label{alg1_step2}
  Initialize $\Vtheta^0$ randomly. 
Set $\hat\Vtheta= \Vtheta^0, \hat{L}= + \infty$. 
  \WHILE{$k<K$ and $\phi < \Phi$} \label{while_loop_1}
  \STATE \label{step_grad_update}
         $\Vtheta^{k+1} = \Vtheta^{k} - {\eta}\nabla_{\Vtheta^{k}}L_{\mathcal{D^{\textrm{train}}}}\big(y;f(\Vx(\hat{\Valpha});\Vtheta^{k})\big)$.
\STATE \label{step_validation_loss} 
Compute ${L}_\text{val} = L_{\mathcal{D}^{\text{val}}}\big(y;f(\Vx(\hat{\Valpha});\Vtheta^{k+1})\big)$.   
\IF{${L}_\text{val} < \hat{L}$}\label{algo1_step6}
    \STATE $\hat\Vtheta = \Vtheta^{k+1}$, 
    $\hat{L} = {L}_\text{val}$, $\phi = 0$; 
\ELSE 
    \STATE \label{algo1_line9} $\phi \leftarrow \phi + 1$.
\ENDIF \label{algo1_line10}
\STATE $k \leftarrow k+1$
\ENDWHILE \label{algo1_line12}
  \end{algorithmic}
\end{algorithm}

Effectively, solving \eqref{nominal_a_model} is equivalent to first observing the missing features and then retraining a model without them, 
which enables full recourse against the realization of $\Valpha$ and exhibits strong performance \cite{tawn2020missing}.
However, regardless if implemented on the fly or offline, retraining would soon become impractical, to say the least,
as the number of possible combinations grows exponentially with the number of features.

\subsection{Robust and Adaptive Formulations of Base Forecasting Models with Missing Data}\label{reg_subsection}

In this work, we model the uncertainty in feature availability
using an uncertainty set defined as follows:
\begin{equation}\label{uncertainty_set}
    \mathcal{U} = \{\Valpha \in \{\mathbf{0}, \mathbf{1}\}^{|\mathcal{P}|} \,|\, \mathbf{1}^{\top}\Valpha \leq \gamma\},
\end{equation}
where $\gamma$ is the maximum number of potentially missing features, 
\emph{a.k.a.} the uncertainty budget, with $\gamma \leq |\mathcal{P}|$.
Next, 
we present robust and adaptive robust formulations for base forecasting models that employ the uncertainty set \eqref{uncertainty_set}.

\subsubsection{Robust Forecasting (RF)}\label{robust_reg_subsection}

A first approach to deal with feature availability uncertainty
is to employ an RF formulation that minimizes the worst-case loss 
when up to $\gamma$ features are missing, given by
\begin{equation}\label{robust_static}
\texttt{RF}:  \quad     {\min}_{\Vtheta}\,
    {\max}_{\Valpha\in \mathcal{U}}\,
{L}_\mathcal{D}\big(y,f(\Vx(\Valpha);\Vtheta)\big),
\end{equation}
where $f$ is the base forecasting model, 
LR or NN, defined by \eqref{LR-base} or \eqref{NN-base}, respectively.
Not surprisingly, 
depending on $\gamma$ and the respective uncertainty set $\mathcal{U}$, \eqref{robust_static} may lead to an overly conservative model parametrization.

\subsubsection{Adaptive Robust Forecasting (ARF)} \label{adapt_LDR_section}

To reduce conservativeness, 
inspired by \cite{bertsimas2024adaptive}, 
we employ an ARF formulation given by
\begin{equation}\label{adaptive_robust_model}
\texttt{ARF}: \quad   {\min}_{\Vtheta(\Valpha)}\,
    {\max}_{\Valpha \in \mathcal{U}}\,
{L}_\mathcal{D}\Big(y,f\big(\Vx(\Valpha);\Vtheta(\Valpha)\big)\Big),
\end{equation}
where the base forecasting model parameters are a function of the uncertain parameters, i.e., $\Vtheta(\Valpha)$.
\black{That is, given a realization of missing features during real-time operations, 
model parameters adapt to the available features.}
Considering linear adaptation, 
i.e., parameters $\Vtheta(\Valpha)$ that are linear in $\Valpha$, 
we obtain the following formulations 
for the base forecasting models.

\paragraph{LR}
For the LR base model \eqref{LR-base}, 
assuming the bias is modeled with a constant feature that equals 1 and is always available, 
we set $\Vtheta(\Valpha) = \Vw + \matD\Valpha$ 
to obtain
\begin{equation*}\label{ARF_lr_model}
    f(\Vx(\Valpha);\Vtheta(\Valpha)) = 
    (\Vw + \matD\Valpha)^\top\Vx(\Valpha).
\end{equation*}
Notably, $\Vw$ represents the model with all features available and $D_{z,j}$ is the linear correction applied to $w_{z}$ when the $j$-th feature is missing.
In the special case of an RF formulation
with piecewise linear loss $l$, 
\eqref{robust_static} can be solved using reformulation-based approaches \cite{stratigakos2023towards}.
However, in energy forecasting tasks, the number of observations $n$ can be very large 
(e.g., in the order of $10^4$ for hourly resolution series),
which may challenge the scalability of such approaches.

\begin{figure}[tb!]
    \centering
    \resizebox{.75\columnwidth}{!}{
    \tikzset{every picture/.style={line width=0.75pt}} 

\begin{tikzpicture}[x=0.75pt,y=0.75pt,yscale=-1,xscale=1]

\draw    (70,50) -- (117,50) ;
\draw [shift={(120,50)}, rotate = 180] [fill={rgb, 255:red, 0; green, 0; blue, 0 }  ][line width=0.08]  [draw opacity=0] (8.93,-4.29) -- (0,0) -- (8.93,4.29) -- cycle    ;
\draw   (119.5,30) -- (149.5,30) -- (149.5,70) -- (119.5,70) -- cycle ;

\draw   (199.5,30) -- (229.5,30) -- (229.5,70) -- (199.5,70) -- cycle ;
\draw    (229.5,50) -- (256.5,50) ;
\draw [shift={(259.5,50)}, rotate = 180] [fill={rgb, 255:red, 0; green, 0; blue, 0 }  ][line width=0.08]  [draw opacity=0] (8.93,-4.29) -- (0,0) -- (8.93,4.29) -- cycle    ;
\draw    (289.5,50) -- (316.5,50) ;
\draw [shift={(319.5,50)}, rotate = 180] [fill={rgb, 255:red, 0; green, 0; blue, 0 }  ][line width=0.08]  [draw opacity=0] (8.93,-4.29) -- (0,0) -- (8.93,4.29) -- cycle    ;
\draw   (319.5,30) -- (349.5,30) -- (349.5,70) -- (319.5,70) -- cycle ;
\draw    (149.5,50) -- (196.5,50) ;
\draw [shift={(199.5,50)}, rotate = 180] [fill={rgb, 255:red, 0; green, 0; blue, 0 }  ][line width=0.08]  [draw opacity=0] (8.93,-4.29) -- (0,0) -- (8.93,4.29) -- cycle    ;
\draw   (39.5,30) -- (69.5,30) -- (69.5,70) -- (39.5,70) -- cycle ;
\draw   (40,90) -- (70,90) -- (70,130) -- (40,130) -- cycle ;

\draw  [dash pattern={on 4.5pt off 4.5pt}]  (70,100) -- (130,100) -- (130,73) ;
\draw [shift={(130,70)}, rotate = 90] [fill={rgb, 255:red, 0; green, 0; blue, 0 }  ][line width=0.08]  [draw opacity=0] (8.93,-4.29) -- (0,0) -- (8.93,4.29) -- cycle    ;
\draw  [dash pattern={on 4.5pt off 4.5pt}]  (70,110) -- (210,110) -- (210,73) ;
\draw [shift={(210,70)}, rotate = 90] [fill={rgb, 255:red, 0; green, 0; blue, 0 }  ][line width=0.08]  [draw opacity=0] (8.93,-4.29) -- (0,0) -- (8.93,4.29) -- cycle    ;
\draw  [dash pattern={on 4.5pt off 4.5pt}]  (70,120) -- (330,120) -- (330,73) ;
\draw [shift={(330,70)}, rotate = 90] [fill={rgb, 255:red, 0; green, 0; blue, 0 }  ][line width=0.08]  [draw opacity=0] (8.93,-4.29) -- (0,0) -- (8.93,4.29) -- cycle    ;

\draw (207,41) node [anchor=north west][inner sep=0.75pt]   [align=left] {$\displaystyle \mathbf{g}^{2}$};
\draw (265.5,46) node [anchor=north west][inner sep=0.75pt]   [align=left] {$\displaystyle \dotsc $};
\draw (327,41) node [anchor=north west][inner sep=0.75pt]   [align=left] {$\displaystyle f$};
\draw (85,85) node [anchor=north west][inner sep=0.75pt]   [align=left] {$\displaystyle \mathbf{D}^{0}$};
\draw (127,40.83) node [anchor=north west][inner sep=0.75pt]   [align=left] {$\displaystyle \mathbf{g}^{1}$};
\draw (40,41) node [anchor=north west][inner sep=0.75pt]   [align=left] {$\displaystyle \mathbf{x}( \bm{\alpha})$};
\draw (83,36) node [anchor=north west][inner sep=0.75pt]   [align=left] {$\displaystyle \mathbf{W}^{0}$};
\draw (161,36) node [anchor=north west][inner sep=0.75pt]   [align=left] {$\displaystyle \mathbf{W}^{1}$};
\draw (48.5,106.5) node [anchor=north west][inner sep=0.75pt]   [align=left] {$\displaystyle \bm{\alpha}$};
\draw (165,95) node [anchor=north west][inner sep=0.75pt]   [align=left] {$\displaystyle \mathbf{D}^{1}$};
\draw (295,105) node [anchor=north west][inner sep=0.75pt]   [align=left] {$\displaystyle \mathbf{D}^{M}$};

\end{tikzpicture}}
    \caption{A feed-forward neural network with linear correction.}
\label{nn_LDR}
\end{figure}
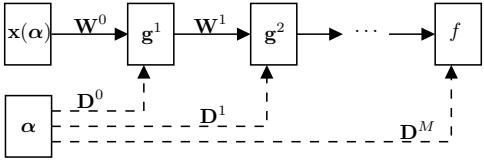

\paragraph{NN}
For the NN base model \eqref{NN-base}, we pass $\Valpha$ onto each layer and apply a linear correction to its weight matrix --- see Fig.~\ref{nn_LDR} for a visualization.
Assuming the respective weight matrices absorb layer biases, 
a linearly adaptive $M$-layer feed-forward NN model is given by
\begin{align*}
    & \mathbf{g}^1 = (\matW^{0} + \matD^{0} \Valpha) \Vx(\Valpha),  \\
    & \mathbf{g}^{m+1} = \max\big(
    (\matW^{m} + \matD^{m} \Valpha)\mathbf{g}^m, \mathbf{0}\big),\,\, 1 \leq m < M, \\    
    & {f}\big(\Vx(\Valpha); \Vtheta(\Valpha)\big) = (\Vw + \matD\Valpha)^{\top} \mathbf{g}^{M}, 
\end{align*}
where $\matD^{m}$ comprises the linear correction terms of the $m$-th hidden layer, and $\Vtheta(\Valpha) = \{(\matW^{0:m-1}+\matD^{0:m-1}\Valpha), 
(\Vw+\matD\Valpha)\}$ denotes the adaptive model parameters.
The number of rows of $\matD^{m}$ equals the number of columns of $\matW^m$ and, with a slight abuse of notation, the matrix-vector addition $\matW^{m} + \matD^{m} \Valpha$ denotes a broadcasting operation, where vector $\matD^{m} \Valpha$ is copied and added to each row of $\matW^{m}$.
Similarly to the LR base model, $\matW^{m}$ denotes the layer weights when all features are available and $D^{m}_{z,j}$ is the linear correction applied to $W^{m}_{z,j}$ when the $j$-th feature is missing.

\subsection{Adversarial Training with Missing Data}\label{adv_learn_methodology_subsection}

The RF \eqref{robust_static} and ARF \eqref{adaptive_robust_model} formulations typically cannot be approximated by a tractable reformulation. 
In what follows,
based on the idea of training with adversarial examples \cite{madry2018towards}, 
we develop a tractable gradient-based algorithm to approximate the solution of \eqref{robust_static} and \eqref{adaptive_robust_model}.
Since the same methodology applies to both RF and ARF formulations, 
we focus our exposition on the former. 
In brief, adversarial training involves:
\emph{(i)} initializing model parameters $\Vtheta$, 
\emph{(ii)} finding an (approximately) optimal $\Valpha$ for the inner maximization problem (adversarial example), 
\emph{(iii)} implementing a gradient update of $\Vtheta$ at the optimum of the inner problem using Danskin's theorem \cite{danskin2012theory}, 
and \emph{(iv)} repeating steps \emph{(ii)}-\emph{(iii)} until convergence.
We note that Danskin's theorem holds for the unique optimum of the inner maximization problem, which we may not be able to find.
\black{Indeed, in the following, we describe a tailored algorithm that approximately solves this inner maximization problem.}
\black{In practice, however, training with approximately optimal adversarial examples provides useful gradients regardless \cite{madry2018towards}.}

Given a fixed set of parameters $\hat\Vtheta$, an adversarial example $\Valpha^{\text{adv}}$ is given by 
\begin{equation}\label{adversarial_example2}
        \Valpha^{\text{adv}} \in 
    {\argmax}_{\Valpha \in \mathcal{U}}\,L_{\mathcal{D}}\big(y;f(\Vx(\Valpha);\hat\Vtheta)\big),
\end{equation}
which is typically some norm maximization problem, non-convex, and thus hard to solve.
Previous energy forecasting applications \cite{wangkun2023, chen2019exploiting} addressed this problem, 
for $l = (\cdot)^2$ (mean squared error loss), 
by finding adversarial examples that strictly minimize or maximize the resultant forecast.
\black{Note that \eqref{adversarial_example2} can be optimally solved by searching over all the potential missing feature combinations, which, however, quickly becomes impractical as the number of features increases.}
Inspired by well-known feature importance metrics, we develop a greedy search algorithm \black{to rapidly and approximately solve \eqref{adversarial_example2} and find useful adversarial examples}, 
detailed in Algorithm~\ref{adv_example_algo}.

Algorithm~\ref{adv_example_algo} uses as input a data set $\mathcal{D}'$,
a feature set $\mathcal{P}'$, 
a specific set of parameters $\hat\Vtheta$, and an ``optimistic'' scenario, $\Valpha^{\textrm{opt}}=\boldsymbol{0}$, in which all features in $\mathcal{P}'$ are available.
It returns an adversarial example $\Valpha^{\text{adv}}$ as follows.
Initially, $\Valpha^{\text{adv}}$ is set equal to $\Valpha^{\textrm{opt}}$,
and a copy of $\mathcal{P}'$, denoted by $\mathcal{C}$, is created (lines~\ref{algo2_step1} and \ref{algo2_step2}).
The empirical loss \eqref{emp_loss} over $\mathcal{D}'$ is evaluated for each potentially missing feature,
setting (temporarily, one by one) the respective entry of $\Valpha^{\text{adv}}$ equal to 1 (lines~\ref{algo2_step4}-\ref{algo2_step7}). 
Then, the feature with the largest loss, $j^*$, is selected,
the respective entry of $\alpha_{j^*}^{\textrm{adv}}$ is fixed to 1, 
and $j^*$ is removed from $\mathcal{C}$ 
(lines~\ref{algo2_step8}-\ref{algo2_step10}).
The evaluation is repeated until reaching $\gamma$ missing features (line~\ref{algo2_step3}) or the loss stops increasing (lines~\ref{algo2_step9}, \ref{algo2_step12}).

\begin{algorithm}[tb!]
\caption{ \texttt{Adversarial Example}}\label{adv_example_algo}
  \textbf{Input:} Data set 
  $\mathcal{D}'$, feature set $\mathcal{P}'$, 
  model parameters $\hat\Vtheta$, $\Valpha^\text{opt}$.  
  \\
  \textbf{Output:} Adversarial $\Valpha^\text{adv}$.
  
  \begin{algorithmic}[1]
  \STATE \label{algo2_step1} Set $\Valpha^\text{adv}=\Valpha=\Valpha^{\textrm{opt}}$, 
   $L^\text{adv} = L_{\mathcal{D}'}\big(y;f(\Vx({\Valpha^{\textrm{adv}}});\hat\Vtheta)\big)$.
  \STATE \label{algo2_step2} Create $\mathcal{C} = \mathcal{P}'$.
    \WHILE{$\mathbf{1}^{\top}\Valpha^\text{adv} < \gamma$}\label{algo2_step3}
    \FOR{$j \in \mathcal{C}$}\label{algo2_step4} 
  \STATE \label{algo2_step5} Set ${\alpha}_j=1$ and 
  ${\alpha}_{j'}=0$, $\forall j' \in \mathcal{C}, \, j'\neq j$.
  \STATE Compute 
  $L(j) = L_{\mathcal{D}'}\big(y;f(\Vx({\Valpha});\hat\Vtheta)\big).$
  \ENDFOR \label{algo2_step7}
  \STATE \label{algo2_step8} Select $j^* \in {\argmax}_{j \in \mathcal{C}}~L(j)$.
  \IF{$L(j^*) \geq L^\text{adv}$}\label{algo2_step9}
    \STATE \label{algo2_step10} set ${\alpha}^\text{adv}_{j^*} = 1$, 
  $L^\text{adv} = L(j^*)$, remove $j^*$ from $\mathcal{C}$
  \ELSE \label{algo2_step11}
  \STATE break. \label{algo2_step12}
  \ENDIF
  \ENDWHILE
  \end{algorithmic}
\end{algorithm}

Algorithm~\ref{adv_train_algo} details
the full adversarial training process to solve \eqref{robust_static} and learn parameters $\Vtheta^{\textrm{adv}}$.
In a nutshell, Algorithm~\ref{adv_train_algo} modifies Algorithm~\ref{nom_train_algo} by introducing adversarial examples found with Algorithm~\ref{adv_example_algo}.
Specifically,
it
initially uses Algorithm~\ref{nom_train_algo} for $\hat\Valpha=\Valpha^{\textrm{opt}}$ to derive 
 $\Vtheta^{\textrm{opt}}$,
 an ``optimistic'' set of parameters 
that optimizes performance for non-adversarial inputs, 
 and sets $\Vtheta^{\textrm{adv}}$ equal to $\Vtheta^{\textrm{opt}}$  
(line~\ref{algo3_step2}).
At iteration $k$, it finds an adversarial example $\Valpha^{\textrm{adv}}$ using Algorithm~\ref{adv_example_algo} in the training data set (line~\ref{algo3_step4})
and performs a gradient update based on the training loss (line~\ref{algo3_step5}).
It then finds a new adversarial example 
using Algorithm~\ref{adv_example_algo} in the validation data set (line ~\ref{algo3_step6}),
and computes the validation loss (line~\ref{algo3_step7});
\black{finding new adversarial examples in the validation data set at each iteration also reduces the overfitting risk.}
Similarly to Algorithm~\ref{nom_train_algo}, 
Algorithm~\ref{adv_train_algo} terminates when 
the maximum number of iterations, $K$, is reached
or the validation loss does not improve for $\Phi$ consecutive iterations.
It returns parameters $\Vtheta^{\textrm{adv}}$, 
and the resulting adversarial loss, $L^{\textrm{adv}}$.

\black{Warm-starting Algorithm~\ref{adv_train_algo} using Algorithm~\ref{nom_train_algo} (line~\ref{algo3_step2}) guarantees that the adversarial loss improves upon the optimistic parameterization. 
In addition, we empirically observed that the warm-start strategy sped up convergence and improved performance under non-adversarial inputs, with the impact being more pronounced for ARF.}

\begin{algorithm}[tb!]
\caption{ \texttt{Adversarial Training Process}}\label{adv_train_algo}
  \textbf{Input:} Data set 
  $\mathcal{D}$, feature set $\mathcal{P}'$,  
  learning rate $\eta$, number of iterations $K$, patience threshold $\Phi$, $\Valpha^\text{opt}$.
  \\
  \textbf{Output:} Parameters $\Vtheta^{\text{adv}}$, loss $L^{\text{adv}}$.
  
  \begin{algorithmic}[1]
\STATE \label{step_initial} 
  Set $k = 0$,  $\phi=0$. Split $\mathcal{D}=\mathcal{D}^{\text{train}}\cup\mathcal{D}^{\text{val}}$.
\STATE \label{algo3_step2} 
Run Algorithm~\ref{nom_train_algo} for $\hat{\Valpha} = \Valpha^{\textrm{opt}}$, and set $\Vtheta^{\text{adv}}= \Vtheta^0 = \Vtheta^{\textrm{opt}}, L^{\text{adv}}= + \infty$.
\WHILE{$k<K$ and $\phi < \Phi$}
\STATE \label{algo3_step4} Run Algorithm~\ref{adv_example_algo} for $\mathcal{D^{\textrm{train}}}$, $\mathcal{P}'$, $\Vtheta^k$, $\Valpha^\text{opt}$, 
to get $\Valpha^{\text{adv}}$.
\STATE \label{algo3_step5} 
$\Vtheta^{k+1} = \Vtheta^{k} - {\eta}\nabla_{\Vtheta^{k}}L_{\mathcal{D^{\textrm{train}}}}\big(y;f(\Vx(\Valpha^{\text{adv}});\Vtheta^{k})\big)$.
         
\STATE \label{algo3_step6} Run Algorithm~\ref{adv_example_algo} for  $\mathcal{D^{\textrm{val}}}$, $\mathcal{P}'$, $\Vtheta^{k+1}$, $\Valpha^\text{opt}$, 
to get $\Valpha^{\text{adv}}$.
\STATE \label{algo3_step7} 
Compute ${L}_\text{val} = L_{\mathcal{D}^{\text{val}}}\big(y;f(\Vx(\Valpha^{\textrm{adv}});\Vtheta^{k+1})\big)$.   
\IF{${L}_\text{val} < L^{\text{adv}}$}
    \STATE $\Vtheta^{\text{adv}} = \Vtheta^{k+1}$, 
    $L^{\text{adv}} = {L}_\text{val}$, 
    $\phi = 0$; 
\ELSE 
\STATE $\phi \leftarrow \phi + 1$.
\ENDIF
    \STATE $k \leftarrow k+1$. 
\ENDWHILE
  \end{algorithmic}
\end{algorithm}

\section{Uncertainty Set Partitioning}\label{partitioning_section}

In this section, we discuss the adaptation to missing data via uncertainty set partitioning (Subsection \ref{set_partion_section}) and propose an algorithm to learn partitions from data (Subsection \ref{learning_partitions_subsection}).

\subsection{Missing Data Adaptation via Uncertainty Set Partitioning}\label{set_partion_section}

Leveraging finite adaptability \cite{bertsimas2010finite, postek2016multistage, stratigakos2023interpretable}, 
we enhance the model capacity to adapt to missing data by partitioning the uncertainty set $\mathcal{U}$ into smaller, hence less conservative, sets.
Let $\mathcal{T} = \{\mathcal{U}_q\}_{q \in [Q]}$  
denote a partition of the uncertainty set $\mathcal{U}$ into $Q$ disjoint subsets, 
such that $\mathcal{U} = \bigcup_{q\in[Q]} \mathcal{U}_q$.
For each subset, $\mathcal{U}_q$, 
we solve \eqref{robust_static} and \eqref{adaptive_robust_model},
to learn a set of model parameters for the RF and ARF formulation, respectively.
Indeed, any realization of $\Valpha$ would fall into exactly one subset.
Hence, when the forecasting model is deployed, 
i.e., $\Valpha$ is realized,
we can identify the partition for the specific realization 
and use the respective model parameters. 
\black{Effectively, this translates into model parameters becoming a piecewise constant (for the RF case) or piecewise linear (for the ARF case) function of $\Valpha$.}
In the extreme case of partitioning $\mathcal{U}$ at each possible missing feature combination, i.e., $Q=|\mathcal{U}|$, 
the outcome would be equivalent to retraining without the missing features \cite{tawn2020missing}.

Since the number of subsets, $Q$, controls the model complexity, 
the critical question is how to partition $\mathcal{U}$.
Previous works \cite{stratigakos2023towards, wangkun2023} split $\mathcal{U}$ using a collection of equality-constrained uncertainty sets with increasing budget.
Namely, \eqref{uncertainty_set} is partitioned at $\gamma+1$ subsets given by
\begin{equation}\label{fixed_partitions}
    \mathcal{U}^{\textrm{fixed}}_{\ell} = \{\Valpha \in \{\mathbf{0}, \mathbf{1}\}^{|\mathcal{P}|} \,|\, \mathbf{1}^{\top}\Valpha = \ell \}, \,\ell \in \{0, 1, \dots, \gamma \},
\end{equation}
i.e., considering each integer value in the range $[0, \gamma]$. 
We refer to \eqref{fixed_partitions} as \textit{fixed} partitioning, since it depends solely on the uncertainty budget and does not consider the available data or the forecasting problem.
However, fixed partitioning is ad-hoc and may be overly conservative. 
We posit that data-driven partitioning can jointly improve the loss for adversarial and non-adversarial examples while requiring fewer subsets.

\subsection{Data-driven Uncertainty Set Partitions} \label{learning_partitions_subsection}

In this subsection, we propose a novel tree-based algorithm to learn a data-driven partition $\mathcal{T}$,
which, in contrast to fixed partitioning, 
leverages available data and tailors the learned partition to the forecasting model.
\black{The algorithm exploits the fact that several combinations of missing data may have a similar loss and thus can be grouped into a subset of combinations.}
For ease of exposition, we present the approach for the RF formulation; 
the application to the ARF formulation is straightforward.

Broadly, given an uncertainty set, say $\mathcal{U}_{\ell}$, 
we estimate a lower bound on the empirical loss, $\texttt{LB}_{\ell}$, 
running Algorithm~\ref{nom_train_algo} for the ``optimistic'' scenario $\Valpha^{\textrm{opt}}_{\ell}$, 
and an upper bound, $\texttt{UB}_{\ell}$, 
running Algorithm~\ref{adv_train_algo} to derive the adversarial loss.
We then search over all the potentially missing features in $\mathcal{P}_{\ell}$, 
select the one that minimizes the relative gap
\begin{equation*}
   \texttt{RelGap}(\mathcal{U}_{\ell}) = \frac{\texttt{UB}_{\ell}-\texttt{LB}_{\ell}}{\texttt{LB}_{\ell}},
\end{equation*}
and split the uncertainty set into two subsets based on the availability of the selected feature.
\black{Effectively, the algorithm seeks a (relatively) small number of critical missing feature combinations and learns dedicated (or ``optimistic'') parameters for each, 
whereas the remaining, less important, combinations are grouped and addressed through adversarial training.}
\black{In the extreme case where $\mathcal{U}_\ell$ contains a single missing feature combination, it follows that $\texttt{RelGap}(\mathcal{U}_{\ell})$ becomes zero.}

Algorithm~\ref{node_split_algo} details the partition process,
which requires controlling for the number of subsets, $Q$, 
and an upper bound on the maximum relative gap, $\varepsilon$, across all subsets, \black{which serves as an early termination criterion}.
The partition is initialized at the original uncertainty set \eqref{uncertainty_set}, denoted by $\mathcal{U}_0$ (line~\ref{algo4_step1}),
for which $\texttt{RelGap}(\mathcal{U}_{0})$, $\Vtheta_0^{\textrm{opt}},$ and $\Vtheta_0^{\textrm{adv}}$ are computed (line~\ref{algo4_step2}).
For the current partition $\mathcal{T}$, 
the subset with the largest relative gap, $\mathcal{U}_{\ell}$, 
is first selected (line~\ref{algo4_step4}),
followed by the feature that leads to the largest loss, $j^*$, 
evaluated using Algorithm~\ref{adv_example_algo},
iterating over the potentially missing features in the set $\mathcal{P}_{\ell}$, with $\Vtheta=\Vtheta^{\textrm{opt}}_{\ell}$ (line~\ref{algo4_step5}).
Set $\mathcal{U}_{\ell}$ is then split into 
$\mathcal{U}_{|\mathcal{T}|+1}$ and 
$\mathcal{U}_{|\mathcal{T}|+2}$, 
by appending equality constraints 
$a_{j^*} = 0$ and $a_{j^*} = 1$, respectively,
and the respective optimistic scenarios $\Valpha_{|\mathcal{T}|+1}^{\textrm{opt}}, \Valpha_{|\mathcal{T}|+2}^{\textrm{opt}}$ are updated
 (line~\ref{algo4_step6}).
For the new subsets, the relative gaps and optimistic/adversarial model parameters are obtained using Algorithms~\ref{nom_train_algo} and \ref{adv_train_algo} (lines~\ref{algo4_step7} and \ref{algo4_step8}).
Since $\mathcal{U}_{|\mathcal{T}|+1}$ fixes a feature as always available, $\texttt{LB}_{|\mathcal{T}|+1}=\texttt{LB}_{\ell}$, 
whereas $\texttt{UB}_{|\mathcal{T}|+1} < \texttt{UB}_{\ell}$.
Similarly, since $\mathcal{U}_{|\mathcal{T}|+2}$ fixes the same feature as always missing,
$\texttt{LB}_{|\mathcal{T}|+2}>\texttt{LB}_{\ell}$, 
whereas
$\texttt{UB}_{|\mathcal{T}|+2}=\texttt{UB}_{\ell}$.
Once the partition is updated (line~\ref{algo4_step9}),
the process (lines~\ref{algo4_step4}-\ref{algo4_step9}) is repeated 
until $Q$ subsets are created or until the maximum relative gap, 
${\max}_{\mathcal{U}_q \in \mathcal{T}} \texttt{RelGap}(\mathcal{U}_q)$, becomes smaller than $\varepsilon$ (line~\ref{algo4_step3}).
\black{
In the extreme case where $Q$ is large enough and $\varepsilon=0$, the learned partition
contains all the possible missing feature combinations, 
recovering the retraining approach.}
\black{The overall process of learning the data-driven partitions requires, at most, running Algorithms~\ref{nom_train_algo} and \ref{adv_train_algo} $Q$ times each.}

Algorithm~\ref{tree-algo} learns the data-driven partition, 
for which the forecasting model is trained.
At the time the forecasting model is deployed, 
we observe the realization of the missing features, say $\hat\Valpha$, 
identify the (unique) subset, say $\mathcal{U}_q$, 
which contains this realization, and derive forecasts as follows:
\emph{(i)} in case $\hat\Valpha = \Valpha^{\textrm{opt}}_{q}$,
then we use model parameters $\Vtheta_{q}^{\textrm{opt}}$;
\emph{(ii)} otherwise, we use model parameters $\Vtheta^{\textrm{adv}}_q$.
\black{Identifying the subset $\mathcal{U}_q$ requires searching over a list of size $Q$, hence, the computational overhead to derive forecasts is minimal.}

\begin{algorithm}[tb!]
\caption{ \texttt{Learn Partition}}\label{tree-algo}
  \textbf{Input:} Data set $\mathcal{D}$, 
  uncertainty set $\mathcal{U}$, 
  number of subsets $Q$, 
  maximum relative gap threshold $\varepsilon$.
  
  \textbf{Output:} Partition $\mathcal{T}$. 
 
  \begin{algorithmic}[1]
    \STATE \label{algo4_step1}
    Set $\mathcal{U}_0 = \mathcal{U}$, 
    $\mathcal{P}_0 = \mathcal{P}$,
    $\mathcal{T} = \{\mathcal{U}_0\}$. 
    \STATE \label{algo4_step2} Compute $\Vtheta^{\textrm{opt}}_0$, $\Vtheta^{\textrm{adv}}_0$, $\texttt{RelGap}(\mathcal{U}_0)$: \\
     (a) Run Algorithm~\ref{nom_train_algo} for $\hat\Valpha=\Valpha_0^{\textrm{opt}}=\mathbf{0}$, to get $\Vtheta^{\textrm{opt}}_0 = \hat{\Vtheta}$, and set $\texttt{LB}_0 = \hat{L}$. \\
    (b) Run Algorithm~\ref{adv_train_algo} for $\mathcal{P}_0$, $\Valpha_0^\text{opt}$, to get $\Vtheta^{\textrm{adv}}_0 = \Vtheta^{\textrm{adv}}$, and set $\texttt{UB}_0 = L^{\text{adv}}$.
    \WHILE{$|\mathcal{T}| < Q$ and ${\max}_{\mathcal{U}_q \in \mathcal{T}} \texttt{RelGap}(\mathcal{U}_q) > \varepsilon$}\label{algo4_step3}
    \STATE \label{algo4_step4}
        Select $\mathcal{U}_{\ell} \in {\argmax}_{\mathcal{U}_q \in \mathcal{T}} \texttt{RelGap}(\mathcal{U}_q)$.
    \STATE \label{algo4_step5}
    Run lines~\ref{algo2_step1}-\ref{algo2_step8} of Algorithm~\ref{adv_example_algo} for $\mathcal{P}_{\ell}$, $\Vtheta^{\textrm{opt}}_{\ell}$, $\Valpha_\ell^\text{opt}$,
    to get feature $j^*$ (to split).
    \STATE \label{algo4_step6} Split $\mathcal{U}_{\ell}$ into $\mathcal{U}_{|\mathcal{T}|+1}$ and $\mathcal{U}_{|\mathcal{T}|+2}$, as follows: \\
    (a) $\quad  \mathcal{U}_{|\mathcal{T}|+1} = \mathcal{U}_{\ell} \cap \{\alpha_{j^*} = 0\}$;\\
    $\,\,\,  \qquad  \mathcal{U}_{|\mathcal{T}|+2} = \mathcal{U}_{\ell} \cap \{\alpha_{j^*} = 1\}$.\\
    (b) Set $\Valpha_{|\mathcal{T}|+1}^\text{opt} =\Valpha_{|\mathcal{T}|+2}^\text{opt} = \Valpha_\ell^\text{opt}$.\\
    \, \quad Update $\alpha_{|\mathcal{T}|+1, j^*}^\text{opt} = 0$, and 
    $\alpha_{|\mathcal{T}|+2, j^*}^\text{opt} = 1$.\\
    (c) Set $\mathcal{P}_{|\mathcal{T}|+1} = \mathcal{P}_{|\mathcal{T}|+2} = \mathcal{P}_\ell\setminus\{j^*\}$.
    \STATE \label{algo4_step7}  Compute $\Vtheta^{\textrm{opt}}_{|\mathcal{T}|+1}, \Vtheta^{\textrm{adv}}_{|\mathcal{T}|+1}, \texttt{RelGap}(\mathcal{U}_{|\mathcal{T}|+1})$: \\
    (a) Set $\Vtheta^{\textrm{opt}}_{|\mathcal{T}|+1}=\Vtheta^{\textrm{opt}}_{\ell}$ and $\texttt{LB}_{|\mathcal{T}|+1}= \texttt{LB}_{\ell}$. \\
    (b) Run Algorithm~\ref{adv_train_algo} for $\mathcal{P}_{|\mathcal{T}|+1}$, $\Valpha_{|\mathcal{T}|+1}^\text{opt}$, to get $\Vtheta^{\textrm{adv}}_{|\mathcal{T}|+1} = \Vtheta^{\textrm{adv}} $, and set $\texttt{UB}_{|\mathcal{T}|+1}=L^{\text{adv}}$.  
    \STATE \label{algo4_step8}  Compute $\Vtheta^{\textrm{opt}}_{|\mathcal{T}|+2}, \Vtheta^{\textrm{adv}}_{|\mathcal{T}|+2}, \texttt{RelGap}(\mathcal{U}_{|\mathcal{T}|+2})$:\\
    (a) Run Algorithm~\ref{nom_train_algo} for $\hat\Valpha=\Valpha_{|\mathcal{T}|+2}^{\textrm{opt}}$, to get $\Vtheta_{|\mathcal{T}|+2}^{\textrm{opt}} = \hat\Vtheta$ and set  $\texttt{LB}_{|\mathcal{T}|+2}=\hat{L}$.\\ 
    (b) Set $\Vtheta^{\textrm{adv}}_{|\mathcal{T}|+2}=\Vtheta^{\textrm{adv}}_{\ell}$ and $\texttt{UB}_{|\mathcal{T}|+2} = \texttt{UB}_{\ell}$.
    \STATE \label{algo4_step9} Update partition $\mathcal{T} \leftarrow \{\mathcal{T}\setminus{\mathcal{U}_{\ell}}\}\cup\{\mathcal{U}_{|\mathcal{T}|+1}, \mathcal{U}_{|\mathcal{T}|+2}\}$.
    \ENDWHILE \label{algo4_step10}
\end{algorithmic}\label{node_split_algo}
\end{algorithm}

\begin{figure}[tb!]
    \centering
    \resizebox{\columnwidth}{!}{
    \usetikzlibrary{matrix, positioning, arrows.meta}



\begin{tikzpicture}[x=0.75pt,y=0.75pt,xscale=1,
  node distance=1cm and 2cm,
  every node/.style={font=\normalsize},
  array/.style={draw, rounded corners, fill=blue!10, inner sep=5pt},
  decision/.style={font=\normalsize, sloped, above},
  arrow/.style={-{Latex[round]}, thick},
  lossbox/.style={draw, fill=orange!50, minimum width=5mm, minimum height=2cm},
  gainbox/.style={draw, fill=green!50, minimum width=5mm, minimum height=1cm},
  gaparrow/.style={-Latex[round], double distance=1mm, draw=black!60},
  level 1/.style={sibling distance=30mm},
  level 2/.style={sibling distance=20mm},
]

\node[array] (u0) {
  \begin{tabular}{cc}
    (0,0,0) & (1,0,0) \\
    (0,0,1) & (1,0,1)\\
    (0,1,0) & (1,1,0)\\
    (0,1,1) & (1,1,1)\\
  \end{tabular}
};

\node[array, below left=of u0] (u1) {
  \begin{tabular}{c}
    (0,0,0) \\
    (0,0,1) \\
    (0,1,0) \\
    (0,1,1)
  \end{tabular}
};

\node[array, below right=of u0] (u2) {
  \begin{tabular}{c}
    (1,0,0) \\
    (1,0,1) \\
    (1,1,0) \\
    (1,1,1)
  \end{tabular}
};

\node[array, below left=of u2] (u3) {
  \begin{tabular}{c}
    (1,0,0) \\
    (1,0,1)
  \end{tabular}
};

\node[array, below right=of u2] (u4) {
  \begin{tabular}{c}
    (1,1,0) \\
    (1,1,1)
  \end{tabular}
};

\draw[arrow] (u0) -- node[decision] {$a_1 = 0$} (u1);
\draw[arrow] (u0) -- node[decision] {$a_1 = 1$} (u2);
\draw[arrow] (u2) -- node[decision] {$a_2 = 0$} (u3);
\draw[arrow] (u2) -- node[decision] {$a_2 = 1$} (u4);

\node[above=0.1cm of u0] {$\mathcal{U}_0$};
\node[above=0.1cm of u1] {$\mathcal{U}_1$};
\node[above=0.1cm of u2] {$\mathcal{U}_2$};
\node[above=0.1cm of u3] {$\mathcal{U}_3$};
\node[above=0.1cm of u4] {$\mathcal{U}_4$};

\begin{scope}[shift={(2.5cm, 0)}, yscale=-1] 
    \tikzset{every picture/.style={line width=0.75pt}} 

    \draw  [fill={rgb, 255:red, 126; green, 211; blue, 33 }  ,fill opacity=1 ][line width=1.5]  (0,0) -- (20,0) -- (20,20) -- (0,20) -- cycle ;
    \draw  [fill={rgb, 255:red, 245; green, 166; blue, 35 }  ,fill opacity=1 ][line width=1.5]  (-30,-80) -- (-10,-80) -- (-10,20) -- (-30,20) -- cycle ;
    \draw   (10,-80) -- (25,-60) -- (18.75,-60) -- (18.75,-20) -- (25,-20) -- (10,0) -- (-2.5,-20) -- (3.75,-20) -- (3.75,-60) -- (-2.5,-60) -- cycle ;

    \draw (-30,22) node [anchor=north west][inner sep=0.75pt]   [align=left] {$\texttt{UB}_{0}$};
    \draw (0,22) node [anchor=north west][inner sep=0.75pt]   [align=left] {$\texttt{LB}_{0}$};
    \draw (30,-40) node [anchor=north west][inner sep=0.75pt]   [align=left] {Gap};   
\end{scope}

\begin{scope}[shift={(6.25cm, -1.75cm)}, yscale=-1] 
    \tikzset{every picture/.style={line width=0.75pt}}        

    \draw  [fill={rgb, 255:red, 126; green, 211; blue, 33 }  ,fill opacity=1 ][dash pattern={on 5.63pt off 4.5pt}][line width=1.5]  (0,0) -- (20,0) -- (20,58) -- (0,58) -- cycle ;
    \draw  [fill={rgb, 255:red, 245; green, 166; blue, 35 }  ,fill opacity=1 ][line width=1.5]  (-30,-40) -- (-10,-40) -- (-10,58) -- (-30,58) -- cycle ;

    \draw (-30,60) node [anchor=north west][inner sep=0.75pt]   [align=left] {$\displaystyle \texttt{UB}_{2}$};
    \draw (0,60) node [anchor=north west][inner sep=0.75pt]   [align=left] {$\displaystyle \texttt{LB}_{2}$};
\end{scope}

\begin{scope}[shift={(-6cm, -2.75cm)}, yscale=-1] 
    \tikzset{every picture/.style={line width=0.75pt}}        

    \draw  [fill={rgb, 255:red, 126; green, 211; blue, 33 }  ,fill opacity=1 ][line width=1.5]  (0,0) -- (20,0) -- (20,20) -- (0,20) -- cycle ;
    \draw  [fill={rgb, 255:red, 245; green, 166; blue, 35 }  ,fill opacity=1 ][dash pattern={on 1.69pt off 2.76pt}][line width=1.5]  (-30,-10) -- (-10,-10) -- (-10,20) -- (-30,20) -- cycle ;

    \draw (-30,22) node [anchor=north west][inner sep=0.75pt]   [align=left] {$\displaystyle \texttt{UB}_{1}$};
    \draw (0,22) node [anchor=north west][inner sep=0.75pt]   [align=left] {$\displaystyle \texttt{LB}_{1}$$ $};
\end{scope}

\begin{scope}[shift={(-1cm, -4.5cm)}, yscale=-1] 
    \draw  [fill={rgb, 255:red, 126; green, 211; blue, 33 }  ,fill opacity=1 ][dash pattern={on 5.63pt off 4.5pt}][line width=1.5]  (0,0) -- (20,0) -- (20,58) -- (0,58) -- cycle ;
    \draw  [fill={rgb, 255:red, 245; green, 166; blue, 35 }  ,fill opacity=1 ][dash pattern={on 5.63pt off 4.5pt}][line width=1.5]  (-30,-10) -- (-10,-10) -- (-10,58) -- (-30,58) -- cycle ;

    \draw (-30,60) node [anchor=north west][inner sep=0.75pt]   [align=left] {$\displaystyle \texttt{UB}_{3}$};
    \draw (0,60) node [anchor=north west][inner sep=0.75pt]   [align=left] {$\displaystyle \texttt{LB}_{3}$};
\end{scope}

\begin{scope}[shift={(10cm, -4.5cm)}, yscale=-1] 
    \draw  [fill={rgb, 255:red, 126; green, 211; blue, 33 }  ,fill opacity=1 ][dash pattern={on 1.69pt off 2.76pt}][line width=1.5]  (0,-30) -- (20,-30) -- (20,58) -- (0,58) -- cycle ;
    \draw  [fill={rgb, 255:red, 245; green, 166; blue, 35 }  ,fill opacity=1 ][line width=1.5]  (-30,-40) -- (-10,-40) -- (-10,58) -- (-30,58) -- cycle ;

    
    \draw (-30,60) node [anchor=north west][inner sep=0.75pt]   [align=left] {$\displaystyle \texttt{UB}_{4}$};
    \draw (0,60) node [anchor=north west][inner sep=0.75pt]   [align=left] {$\displaystyle \texttt{LB}_{4}$};
\end{scope}

\end{tikzpicture}

}
    \caption{Partitioning of set $\mathcal{U}_0$.
    Orange and green bars indicate the subset's lower (\texttt{LB}) and upper (\texttt{UB}) bound.
    The bar outline indicates the same height.}
\label{node_split}
\end{figure}
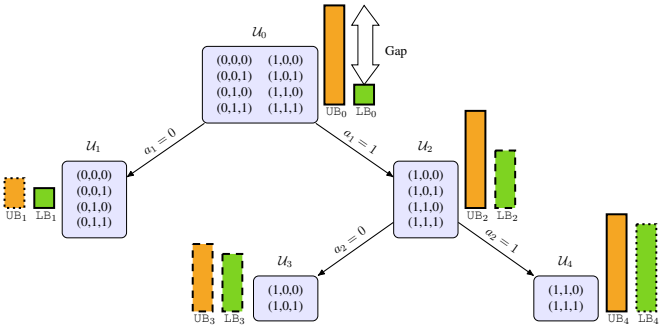

Figure~\ref{node_split} shows an illustrative example, considering a model with up to 3 missing features, i.e., 8 potential combinations of missing features.
The initial uncertainty set $\mathcal{U}_0$ is split on feature $\alpha_1$ 
to derive subsets $\mathcal{U}_1= \mathcal{U}_0 \cap \{\alpha_{1} = 0\} $ and $\mathcal{U}_2 = \mathcal{U}_0 \cap \{\alpha_{1} = 1\}$, 
with the respective index sets of potentially missing features being $\mathcal{P}_1, \mathcal{P}_2=\{2,3\}$ and the ``optimistic'' scenarios being $\Valpha^{\textrm{opt}}_1=(0,0,0)$ and $\Valpha^{\textrm{opt}}_2=(1,0,0)$.
For $\mathcal{U}_1$, the lower bound remains the same ($\texttt{LB}_1=\texttt{LB}_0$), 
whereas the upper bound improves ($\texttt{UB}_1<\texttt{UB}_0$).
For $\mathcal{U}_2$, the upper bound remains the same ($\texttt{UB}_2=\texttt{UB}_0$), 
whereas the lower bound improves ($\texttt{LB}_2 > \texttt{LB}_0$).
For this example, we assume 
$\texttt{RelGap}(\mathcal{U}_1) <  \varepsilon$ and $\texttt{RelGap}(\mathcal{U}_2) > \varepsilon$, hence $\mathcal{U}_2$ is split further.
The final partition is $\mathcal{T} = \{{\mathcal{U}_1, \mathcal{U}_3, \mathcal{U}_4}\}$.

\section{Experimental Setup}\label{exp_setup_section}

In this section, we detail the data and the forecast setting of our numerical experiments on short-term wind power forecasting (Subsection~\ref{data_subsection}), 
benchmark performance without missing data (Subsection~\ref{benchmark_subsection}),
and present the forecasting models with missing data (Subsection~\ref{subsection_missing_data_methods}).

\subsection{Data and Forecast Setting}\label{data_subsection}

We use wind power measurements with a 15-minute granularity 
from $8$ adjacent wind power plants with individual nominal capacities ranging from $80$ to $300\,\textrm{MW}$ and a total nominal capacity of $1127.45\,\textrm{MW}$ \cite{nrel2024dataset}.
The wind power plants are located in the North load zone of the NYISO balancing area, spanning approximately a $55 \times 80\, \textrm{km}$ area of semi-complex terrain.
The data set spans 2018 and we apply a $50/50$ training/test split, with the last $15\%$ of the training data set reserved for validation and hyperparameter tuning.

We examine forecast horizons for $h=\{1, 4, 8, 16\}$ steps ahead, 
i.e., from 15 minutes to 4 hours ahead, 
which are used in short-term scheduling applications of system operators for reliability purposes.
A separate forecasting model is trained for each forecast horizon.
We considered measurements for periods $t$, $t-1$, and $t-2$, for the $8$ plants as features for $8\times3=24$ features.
To account for the weather impact,
which becomes particularly relevant as the horizon increases \cite{alessandrini2017forecast_errors}, 
we considered an additional feature of an hourly production forecast,
issued from \cite{nrel2021reV} 6 hours earlier (at period $t-24$) using the latest available weather forecasts.
We considered that the feature is always available operationally, as these forecasts are issued in batches.
Henceforth, missing data refer to the wind plant production measurements, i.e., $|\mathcal{P}|=24$. 

\subsection{Benchmarks without Missing Data}\label{benchmark_subsection}

We consider the following base forecasting models:
\begin{itemize}
\item \texttt{Pers}, a persistence forecasting model,
industry's standard for very short-term wind power forecasting,
where the last measured value is used as forecast;
\item  \texttt{LR} model \eqref{LR-base} with loss function $l=(\cdot)^2$ (mean squared error), and
\item \texttt{NN} model \eqref{NN-base}, with the same loss function and $4$ hidden layers of $50$ nodes each.
\end{itemize}

We implemented a mini-batch version of Algorithm~\ref{nom_train_algo} with 
batch size $512$, 
learning rate $\eta=10^{-3}$ 
adapted dynamically with the Adam optimizer, 
maximum number of iterations $K=1000$, 
and patience threshold $\Phi=20$.
For \texttt{NN}, 
we additionally included weight decay parameter $10^{-5}$, 
which induces a small regularization effect.

Table~\ref{table_nominal_loss_horizons} presents the root mean squared error (RMSE) (without missing data), normalized by actual observations, for a medium-sized wind plant of $100\,\textrm{MW}$ nominal capacity.
The results indicate that
\texttt{Pers} performs the worst with the highest RMSE across all horizons.
For very short-term forecast horizons up to one hour ahead (see $h=1,4$), \texttt{LR} performs best. 
For horizons longer than 2 hours ahead (see $h=8,16$), \texttt{NN} is the best-performing model and its relative improvement over \texttt{LR} increases with the horizon.
Overall, the performance is consistent with state-of-the-art short-term wind power forecasting in semi-complex terrain \cite{alessandrini2017forecast_errors}.

\begin{table}[tb!]
\caption{
RMSE (\%) without missing data for different horizons $h$ (15 minutes). 
Bold font indicates the best model per $h$.}
\centering
\resizebox{0.6\columnwidth}{!}{
\begin{tabular}{@{}lllll@{}}
\toprule
Horizon      & $h = 1$  &  $h = 4$ & $h=8$ & $h = 16$ \\ \midrule
\texttt{Pers}  & {3.51}     & 9.91 & 15.51  & 23.83 \\
\texttt{LR}    & \textbf{2.58}     &   \textbf{7.97}  & {12.12}  & 16.73 \\
\texttt{NN}    & 2.76   & {8.29} & \textbf{11.69} & \textbf{14.74} \\ \bottomrule
\end{tabular}\label{table_nominal_loss_horizons}}
\end{table}

\subsection{Forecasting Models with Missing Data}\label{subsection_missing_data_methods}

Given a base forecasting model (LR or NN), 
we consider the following methods for handling missing data: 
\begin{itemize}
    \item Imputation (\texttt{Imp}) where missing data are imputed using persistence forecasts \black{(i.e., when a measurement is missing, we forward-fill the last known value)}, industry's  standard practice;
    \item \texttt{RF} \eqref{robust_static} where model parameters are optimized for the worst-case realization of missing data, and 
    \item \texttt{ARF} \eqref{adaptive_robust_model} with linear adaptation to the worst-case realization of missing data.
\end{itemize}

We further consider two partitioning approaches (applicable to \texttt{RF} and \texttt{ARF}):
\begin{itemize}
    \item \texttt{fixed} --- see \eqref{fixed_partitions}, where we split $\mathcal{U}$ at each integer value in the range $[0,\gamma]$, with $\gamma = 24$, and
    \item $\texttt{learn}^Q$, where we learn $Q$ partitions using Algorithm~\ref{tree-algo}, setting $Q=10$ and $\varepsilon=0.1\%$.
\end{itemize}

To ease the exposition, we refer to a forecasting model with missing data by first using the base forecasting model, 
followed by the method to handle missing data, and the partitioning approach (if applicable) in parenthesis.
For example:
\begin{itemize}
    \item \texttt{LR-Imp} denotes an LR base forecasting model with imputation;
    \item \texttt{NN-RF-fixed} denotes an NN base forecasting model, 
with an RF formulation \eqref{robust_static} and fixed partitioning;
    \item \texttt{NN-ARF-learn$^{10}$} denotes an NN base forecasting model, 
with an ARF formulation \eqref{adaptive_robust_model} and $Q=10$ learned partitions, and so forth.
\end{itemize}

We implemented a mini-batch version of Algorithm~\ref{adv_train_algo} with the same hyperparameters as Algorithm~\ref{nom_train_algo}.
For \texttt{fixed} partitions,
we adapted Algorithm~\ref{adv_example_algo} and constructed adversarial examples for equality-constrained uncertainty sets of the form \eqref{fixed_partitions} by sampling missing features with a uniform probability distribution.

\section{Results}\label{results_section}

In this section, we present forecasting results with missing data (Subsection~\ref{subsection_results_randomly}), 
showcase and interpret the proposed methods (Subsection~\ref{interpretation_subsect}), 
and explore the sensitivity to the number of subsets learned (Subsection~\ref{sensitivity_subsection}).
All results pertain to the same $100\,\textrm{MW}$ wind plant of Table~\ref{table_nominal_loss_horizons}.\footnote{The code for recreating the experiments is available on GitHub: \url{https://github.com/akylasstrat/wind-forecast-missing-data}.}
 
\subsection{Performance with Randomly Missing Data}\label{subsection_results_randomly}

\begin{figure*}[tb!]
\centering
\subfloat[\texttt{LR} base forecasting models.]{\includegraphics[width =7 in]{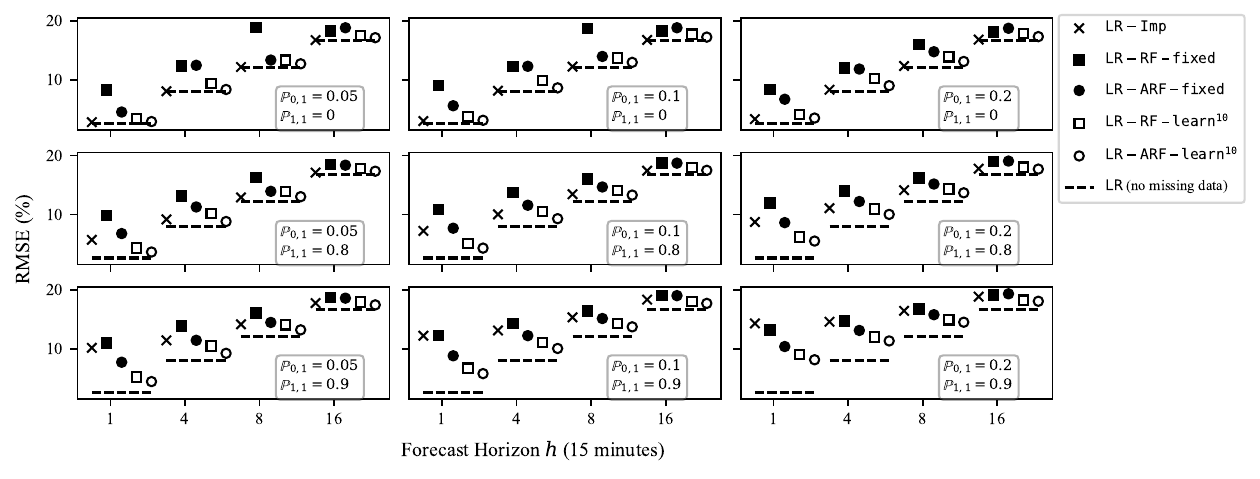}\label{LS_mcar_rmse_horizons}}

\subfloat[\texttt{NN} base forecasting models.]{\includegraphics[width =7 in]{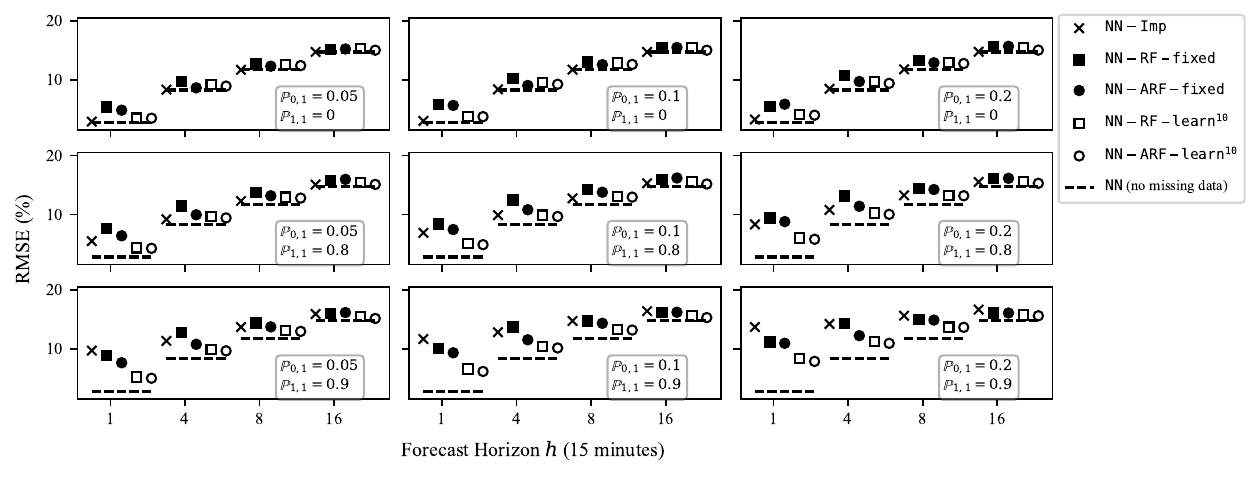}\label{NN_mcar_rmse_horizons}}

\caption{
RMSE (\%) for different forecast horizons with randomly missing data for $\mathbb{P}_{0\rightarrow1}=\{0.05, 0.1, 0.2\}$ and $\mathbb{P}_{1\rightarrow1}=\{0.0, 0.8, 0.9\}$, averaged over $10$ runs.
The dashed line shows performance without missing data.}
\label{LS_NN_joint_mcar_rmse_horizons}
\end{figure*}

We consider randomly missing data, i.e., the probability of missing a measurement is independent of its value, described by a transition matrix.
Let $\mathbb{P}_{0\rightarrow1}$ be the probability that a measurement is missing in a certain period, 
whereas it was available in the previous period.
Further, let $\mathbb{P}_{1\rightarrow1}$ be the probability that a measurement is missing in a certain period, 
whereas it was missing in the previous period.

Figure~\ref{LS_NN_joint_mcar_rmse_horizons} shows the average RMSE over $10$ iterations 
for \texttt{LR} and \texttt{NN} base forecasting models (top and bottom, respectively),
and for different probabilities $\mathbb{P}_{0\rightarrow1}$ and $\mathbb{P}_{1\rightarrow1}$, 
where the probabilities are applied to the $8$ available wind plants.
For each base forecasting model, the $9$ subplots are arranged so that 
each row represents $\mathbb{P}_{1\rightarrow1}$ equal to $0.0, 0.8, 0.9$, 
and each column represents $\mathbb{P}_{0\rightarrow1}$ equal to $0.05, 0.1, 0.2$.
As we move to the right and bottom,
data are missing more frequently ($\mathbb{P}_{0\rightarrow1}$ increases)
and remain unavailable for longer periods 
($\mathbb{P}_{1\rightarrow1}$ increases), respectively.
Indeed, the RMSE for each forecasting model increases
as we move to the right and bottom
as more data are missing.

\subsubsection{\texttt{LR} base forecasting models}
Figure~\ref{LS_mcar_rmse_horizons} illustrates the average RMSE for the \texttt{LR} base forecasting models.
Consider first the imputation-based model $\texttt{LR-Imp}$ ($\times$ marker).
In the top row, where $\mathbb{P}_{1\rightarrow1} = 0$, which implies that the data are missing for a single period,
\texttt{LR-Imp} performs best, 
with an RMSE close to the benchmark value without missing data (dashed line).
As we move to the middle and bottom rows and the probability $\mathbb{P}_{1\rightarrow1}$ increases,
i.e., data are missing for longer periods,
\texttt{LR-Imp} degrades considerably.
The \texttt{LR-ARF-learn$^{10}$} model (empty circle marker) 
performs better in the middle and the bottom row followed closely by \texttt{LR-RF-learn$^{10}$} (empty square marker);
both stay very close to \texttt{LR-Imp} in the top row.
{\black{Pairwise statistical tests \cite{diebold2002comparing} confirmed that \texttt{LR-RF-learn$^{10}$}, 
\texttt{LR-ARF-learn$^{10}$} significantly outperform \texttt{LR-Imp} as $\mathbb{P}_{1\rightarrow1}$ increases.}}
The \texttt{LR-ARF-fixed} and \texttt{LR-RF-fixed} models
perform worse in the top row, 
whereas they become comparable with \texttt{LR-Imp} in the middle row,
and occasionally better in the bottom row, where \texttt{LR-ARF-fixed} generally performs better than \texttt{LR-RF-fixed}.
Considering the performance for different forecast horizons,
we observe that the RMSE difference compared to the benchmark (without missing data) decreases for longer horizons, as evident by the span of the RMSE between the different methods and the benchmark.
For instance, consider the bottom right subplot which is the case with the most missing data;
the RMSE difference of \texttt{LR-ARF-learn$^{10}$} with the benchmark (without missing data)
is about $5.56\%$ for $h=1$,
$3.37\%$ for $h=4$,
$2.39\%$ for $h=8$,
and $1.71\%$ for $h=16$.
In general, \texttt{learned} partitions outperform \texttt{fixed} partitions, but the improvement decreases with the horizon. 
For instance, the best-performing \texttt{LR-ARF-learn$^{10}$} outperforms \texttt{LR-ARF-fixed}
by $39\%, 22\%, 8\%$ and $7\%$, for horizons h equal to $1,4, 8$ and $16$, respectively, averaged across all subplots.
Comparing the \texttt{ARF} with the \texttt{RF} method,
a general observation is that the difference is larger for \texttt{fixed} partitions and shorter horizons,
whereas it decreases for \texttt{learned} partitions and longer horizons.

\subsubsection{\texttt{NN} base forecasting models} Figure~\ref{NN_mcar_rmse_horizons} illustrates the average RMSE for the \texttt{NN} base forecasting models.
Overall, the forecasting models perform similarly to the \texttt{LR} base forecasting models.
The imputation-based model \texttt{NN-Imp} performs the best in the top row, closely followed by \texttt{NN-RF-learn$^{10}$} and \texttt{NN-ARF-learn$^{10}$}.
The latter perform best (and remain close to each other) in the middle and bottom rows.
{\black{Similarly to the \texttt{LR} case, we performed pairwise tests \cite{diebold2002comparing} and confirmed that \texttt{NN-RF-learn$^{10}$}, 
\texttt{NN-ARF-learn$^{10}$} 
provide statistically significant improvement over \texttt{NN-Imp} for larger values of $\mathbb{P}_{1\rightarrow1}$.}}
In general, it is evident that the span between the different models in Fig.~\ref{NN_mcar_rmse_horizons} is quite smaller compared to Fig.~\ref{LS_mcar_rmse_horizons},
and still decreases for longer horizons.
However, the RMSE of the best-performing \texttt{NN-ARF-learn$^{10}$} model is, 
in most cases, higher compared to the respective RMSE of \texttt{LR-ARF-learn$^{10}$} for $h=1,4$ (where \texttt{LR} prevails in Table~\ref{table_nominal_loss_horizons})
whereas it is smaller for $h=8,16$
(where \texttt{NN} prevails in Table~\ref{table_nominal_loss_horizons}).
Similarly to the \texttt{LR} base forecasting models, 
\texttt{learned} partitions outperform \texttt{fixed} partitions and the improvement decreases with the horizon. 
For instance, the best-performing \texttt{NN-ARF-learn$^{10}$} outperforms \texttt{NN-ARF-fixed}
by $32\%, 6\%, 4\%,$and $4\%$, for horizons $h$ equal to $1,4,8,$ and $16$, respectively, averaged across all subplots.
However, the difference between the \texttt{ARF} and \texttt{RF} methods is smaller compared to the \texttt{LR} base models, 
indicating that linear adaptation benefits the latter more.

\subsection{Illustration of the Partitioning and Adaptation Methods}\label{interpretation_subsect}

We illustrate how the proposed methods deal with missing data by examining the learned data-driven partitions and respective model parameterization.

\subsubsection{Initial Splits and Optimistic Weights}
\begin{figure}[tb!]
\centerline{\includegraphics[width =
\columnwidth]{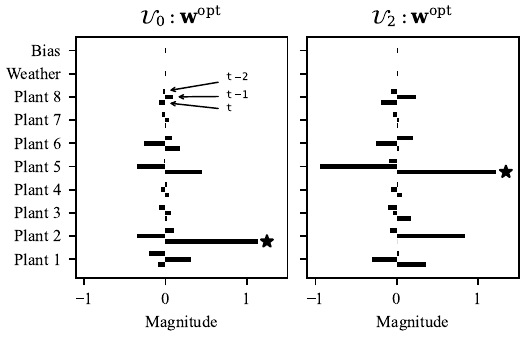}}
\caption{Optimistic weights for subsets $\mathcal{U}_0, \mathcal{U}_2$ for \texttt{LR-ARF-learn} ($h=1$).
The star marker indicates the selected feature selected for a split.
Measurement data are grouped per wind plant, in the order of $t, t-1, t-2$ moving upwards.}
\label{feat_opt_weights}
\end{figure}

Consider the \texttt{LR-ARF-learn$^{Q}$} model for $h=1$.
Table~\ref{tree_structure_table} details the first $2$ splits learned by Algorithm~\ref{tree-algo}.
At $\mathcal{U}_0$, the measurement of plant $2$ (the target plant) at period $t$ is identified as the most important feature ($j^*=3$) and $\mathcal{U}_0$ is split into $\mathcal{U}_1$ ($\alpha_3=0$) and $\mathcal{U}_2$ ($\alpha_3=1$).
$\texttt{RelGap}(\mathcal{U}_1)$ is small ($0.28\%$), which is expected as $\mathcal{U}_1$ is conditioned on the most important feature being available; 
$\mathcal{U}_1$ is not split further.
Conversely, $\texttt{RelGap}(\mathcal{U}_2)$ remains large ($32.42\%$).
Algorithm~\ref{tree-algo} next identifies the measurement of plant $5$ at period $t$ (feature $12$) as the most important feature, 
which is sensible as plant $5$ is adjacent to the target plant and their output are correlated. 
$\mathcal{U}_2$ is then split into $\mathcal{U}_3$ ($\alpha_{12}=0$) and $\mathcal{U}_4$ ($\alpha_{12}=1$), 
inducing the data-driven partition $\mathcal{T}=\{\mathcal{U}_1,\mathcal{U}_3,\mathcal{U}_4\}$ as shown in Table~\ref{tree_structure_table}.
To further understand the induced partition, Fig.~\ref{feat_opt_weights} plots the optimistic weights, $\Vw^{\textrm{opt}}$, for subsets $\mathcal{U}_0$, $\mathcal{U}_2$.
All features, except the bias, represent normalized wind production.
The left subplot shows that the measurement of plant $2$ at period $t$ has the highest absolute weight among all features, which explains the first split,
followed by the measurement of plant $5$ at period $t$.
The latter becomes the most important feature for $\mathcal{U}_2$, as shown in the right subplot, 
which explains the second split.

\begin{table}[tb!]
\caption{First $2$ splits of Algorithm~\ref{tree-algo} for $\texttt{LR-ARF-learn}$ ($h=1$).}
\centering
\resizebox{\columnwidth}{!}{
\begin{tabular}{@{}llllll@{}}
\toprule
Subset & Next Split & $100\cdot\texttt{UB}$ & $100\cdot\texttt{LB}$ & \texttt{RelGap} (\%)  \\ \midrule
$\mathcal{U}_0$ &  Feat. $3$ \, (plant $2$, $t)$ & 5.05 & 0.08 & 61.21 \\
$\mathcal{U}_1$ = $\mathcal{U}_0 \cap \{\alpha_3=0\}$  & -  & 0.28 & 0.08 & 2.40 \\
$\mathcal{U}_2$ = $\mathcal{U}_0 \cap \{\alpha_3=1\}$  & Feat. $12$ (plant $5$, $t$) & 4.71 & 0.14 & 32.42 \\
$\mathcal{U}_3$  =  
$\mathcal{U}_2 \cap \{\alpha_{12}=0\}$   & - & 0.52 & 0.14 & 2.71  \\
$\mathcal{U}_4$ = $\mathcal{U}_2 \cap \{\alpha_{12}=1\}$   & Feat. $4$ \, (plant $5$, $t-1$) & 4.71 & 0.19 & 23.26 \\
\bottomrule
\end{tabular}}\label{tree_structure_table}
\end{table}

\begin{figure}[tb!]
\centerline{\includegraphics[width =
\columnwidth]{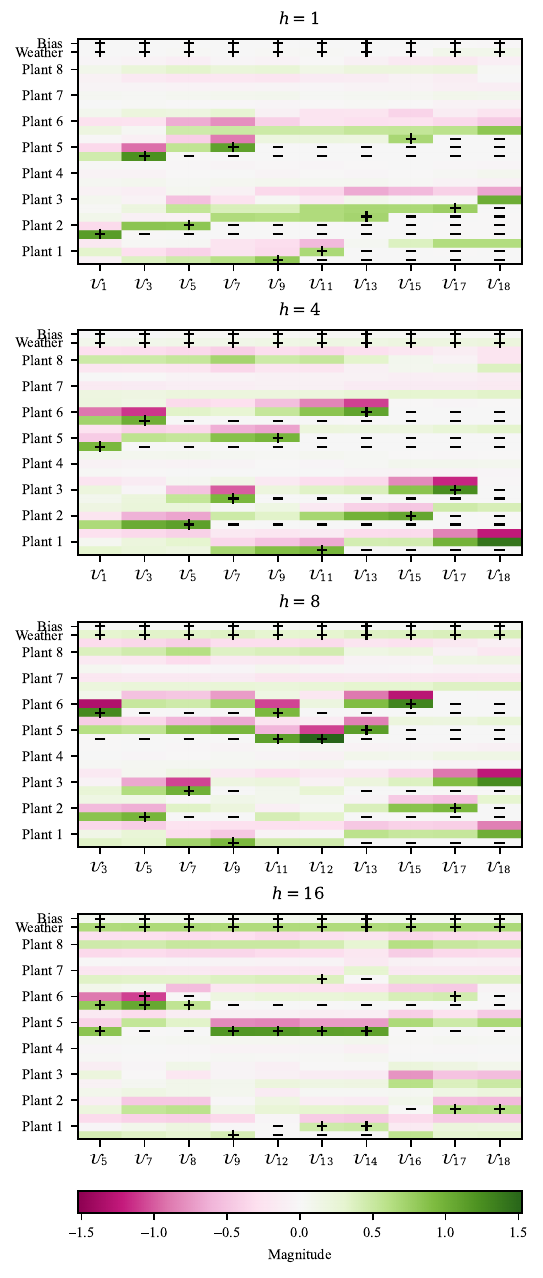}}
\caption{Heatmap of optimistic weights for the final partitions for \texttt{LR-ARF-learn}$^{10}$ for $h=1, 4, 8, 16$.
For each subset, the $+$ marker indicates features fixed as available, whereas the $-$ marker indicates missing features.}
\label{heat_opt_all_weights}
\end{figure}

\begin{figure}[tb!]
\centerline{\includegraphics[width =
\columnwidth]{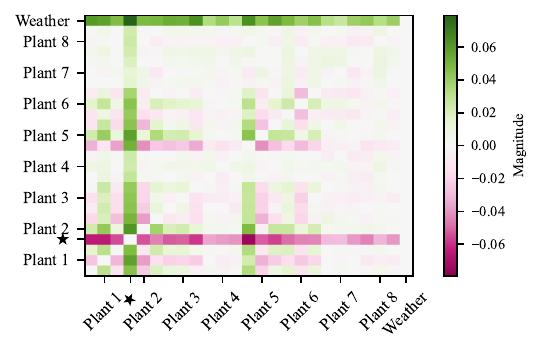}}
\caption{
Linear correction weights $\matD^{\textrm{adv}}$ learned for $\texttt{LR-ARF}$ ($h=1$) for $\mathcal{U}_0$.
The $\bigstar$ marker shows the feature that was subsequently selected for splitting.}
\label{heatmap_linear_correction}
\end{figure}

\begin{figure}[t]
\centerline{\includegraphics[width =
\columnwidth]{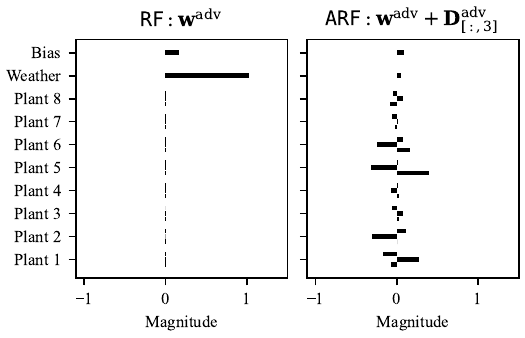}}
\caption{Adversarial weights for $\texttt{LR-RF}$ and $\texttt{LR-ARF}$ ($h=1$) for $\mathcal{U}_0$ conditioned on feature $3$ missing ($\alpha_3=1$).}
\label{feat_adv_weights}
\end{figure}

\subsubsection{Full Partitions for all Horizons}
\black{Figure~\ref{heat_opt_all_weights} illustrates the final partitions and the respective optimistic weights, $\Vw^{\textrm{opt}}$, across all horizons (i.e., $h = 1, 4, 8,16$) for \texttt{LR-ARF-learn$^{10}$}. 
In each subplot, each column represents a subset in the final data-driven partition.
In each subset, the $+$ marker indicates features fixed as available, 
whereas the $-$ marker indicates missing features.
Hence, both markers indicate features that were used for splitting, 
except for weather and bias (top two rows), which are always available.
Moving from the left to the right columns further shows the features that were prioritized during splitting.
For instance, the top subplot ($h=1$) shows that $\mathcal{U}_1=\mathcal{U}_0 \cap \{\alpha_3=0\}$, 
$\mathcal{U}_3=\mathcal{U}_0 \cap \{\alpha_3=1, \alpha_{12}=0\}$ --- also shown in Table~\ref{tree_structure_table} ---, 
$\mathcal{U}_5=\mathcal{U}_0 \cap \{\alpha_3=1, \alpha_{12}=1, \alpha_{4}=0\}$, and so forth.}
\black{Observing the markers and respective feature weights, 
we note that as the forecast horizon increases (moving from the top to the bottom subplot), 
the splits of the partitioning algorithm shift from features of plant $2$ (target plant) to features of plants $5$ and $6$.
Indeed, measurements from adjacent plants tend to become more important than measurements from the target plant for longer horizons, and this feature importance is reflected in the data-driven partitions.}
\black{In addition, for $h=1,4$, the partitioning algorithm prioritizes depth, as every time a feature is set as available, the respective subset is not split further (all columns contain a single $+$ marker, excluding the weather and bias).
Conversely, for $h=8,16$, the (always available) weather feature becomes more important, 
leading to the partitioning algorithm prioritizing breadth (subsets now contain multiple features fixed as available).}
{Similar results are observed for the \texttt{NN} base forecasting models, with initial splits being largely the same as the \texttt{LR} base forecasting models shown in Fig.~\ref{heat_opt_all_weights}.}

\subsubsection{\texttt{RF} vs. \texttt{ARF}}
We now examine the effect of the adaptation methods, i.e., compare \texttt{RF} and \texttt{ARF} formulations.
By design, for a given subset $\mathcal{U}_q$, 
both formulations learn the same optimistic parameters 
(as $\matD^{\textrm{opt}}=\mathbf{0}$) and, hence, 
the same data-driven partitions.
\black{Figure~\ref{heatmap_linear_correction} plots the linear correction weights $\matD^{\textrm{adv}}$ learned for \texttt{LR-ARF} for $\mathcal{U}_0$, 
with each row showing the linear correction applied to the rest of the features when the index feature is missing.
Notably, the top row of Fig.~\ref{heatmap_linear_correction} contains the higher weights, 
meaning that the weather-based feature is used to compensate for missing features.
This is further highlighted in Fig.~\ref{feat_adv_weights}, 
which plots the adversarial weights of $\texttt{LR-RF}$ and $\texttt{LR-ARF}$ for $\mathcal{U}_0$.}
The left subplot shows that $\texttt{RF}$ leads to conservative parameters as it sets all measurement weights close to $0$ (which could be missing) and assigns higher weights to the bias and the weather forecast.
The right subplot shows the adversarial weights of \texttt{ARF}, conditioned on the missing measurement of plant $2$ at period $t$.
The linear corrections $\mathbf{D}^{\textrm{adv}}_{[:,3]}$, \black{highlighted with the $\bigstar$ marker column-wise in Fig.~\ref{heatmap_linear_correction}}, 
counterbalance the nullified $w^{\textrm{adv}}_{3}$, 
which is large and positive, by increasing the rest of the weights;
\black{higher weights are assigned to the weather and measurements of plants $2$ and $5$ at period $t-1$, which are highly correlated with the measurement of plant $2$ at period $t$.}
Overall, as these adversarial weights adapt to the realization of missing features, 
the \texttt{ARF} formulation leads to a less conservative parametrization compared to the \texttt{RF} formulation.

\subsection{Impact of Number of Subsets $Q$}\label{sensitivity_subsection}

\begin{table}[]
\caption{
\texttt{LR-ARF-learn$^{Q}$} performance as a function of $Q$ ($h=1, \mathbb{P}_{0\rightarrow1}=0.2, \mathbb{P}_{1\rightarrow1}=0.9$).}
\centering
\resizebox{0.85\columnwidth}{!}{
\begin{tabular}{@{}llllllll@{}}
\toprule
Q      & 1     & 2     & 5     & 10   & 20   & 50  & 100 \\ \midrule
RMSE (\%)   & 16.06 & 13.03 & 10.54 & 8.14 & 7.51 & 7.45 & 7.43 \\
Max. \texttt{RelGap} (\%) & 61.21 & 32.42 & 17.94 & 6.16 & 2.58 & 1.89 & 1.64 \\
\bottomrule
\end{tabular}}\label{sensitivity_Q_table}
\end{table}

We now examine the impact of the number of subsets $Q$ of the data-driven partitioning algorithm.
Consider the \texttt{LR-ARF-learn$^Q$}  model trained for $h=1$ and evaluated for $\mathbb{P}_{0\rightarrow1} = 0.2, \mathbb{P}_{1\rightarrow1} = 0.9$ --- see bottom right subplot of Fig.~\ref{LS_mcar_rmse_horizons}.
Table~\ref{sensitivity_Q_table} presents the RMSE as a function of $Q$ alongside the maximum relative gap, $\max_{\mathcal{U}_q\in\mathcal{T}}\texttt{RelGap}(\mathcal{U}_q)$.
As expected, increasing $Q$ lowers all metrics in Table~\ref{sensitivity_Q_table}, 
leading to a lower RMSE than \texttt{LR-Imp} for as few as $Q=2$ subsets and $\texttt{LR-ARF-fixed}$ for $Q=5$.
This is intuitive as a small number of features, namely measurements from plants $2$ and adjacent plants, impact accuracy considerably.
The RMSE plateaus for $Q>20$; 
setting $Q=100$ leads to an RMSE difference of $0.08\%$ compared to $Q=20$.
The maximum \texttt{RelGap} is highly correlated with the RMSE, decreasing rapidly for small $Q$ and slowly for larger $Q$.

For longer forecast horizons, we observe that all metrics plateau earlier, which is expected as the impact of missing data and, subsequently, the gains of data-driven partitions are smaller.
For $Q=1$, the maximum \texttt{RelGap} is approximately $5.6\%$ for $h=4$, $2.2\%$ for $h=8$, and $0.83\%$ for $h=16$. 
For \texttt{RelGap} values smaller than $3\%$ the effect on the RSME is in the second decimal.
Overall, the maximum \texttt{RelGap} and its decrease rate correctly identify the potential benefits of further splits.
Therefore, one could assume that if the RMSE, evaluated on a hold-out set, is satisfactory and the maximum \texttt{RelGap} decreases slowly, Algorithm~\ref{tree-algo} could arguably terminate early.

\section{Conclusions}\label{conclusions_section}

Missing data during real-time operations compromises forecast accuracy and leads to suboptimal downstream decisions.
Using ideas from adaptive robust optimization and adversarial machine learning, we developed linear- and neural network-based short-term wind power forecasting models whose parameters seamlessly adapt to the available features.
Our robust and adaptive forecasting models do not rely on historical missing data patterns and are suitable for real-time operations under stringent time constraints.
We conducted extensive numerical experiments on short-term wind power forecasting with horizons ranging from 15 minutes to 4 hours ahead, 
comparing against the industry standard imputation approach.
Our proposed models were on par with imputation when data were missing for very short periods (i.e., when the latest wind power measurement was missing) and significantly better when data were missing for longer periods, hence hedging effectively against worst-case scenarios.

\black{There are several interesting directions for future work,
such as dynamically updating the learned partitions to address potential data non-stationarity, 
and robustifying against adversarial attacks that jointly consider data availability and integrity, e.g., noisy inputs due to measurement errors.}

\bibliographystyle{IEEEtran}
\bibliography{IEEEabrv, ref-short}

\end{document}